\documentclass[10pt]{article} 
\usepackage[accepted]{rlc}

\usepackage{amssymb}            
\usepackage{mathtools}          
\usepackage{mathrsfs}           
\usepackage{graphicx}           
\usepackage{subcaption}
\usepackage{multirow}
\usepackage{booktabs}
\usepackage[space]{grffile}     
\usepackage{url}                

\title{Offline Reinforcement Learning 

with Imputed Rewards}

\author{Carlo Romeo  \\
    carlo.romeo@unifi.it \\
    MICC, University of Florence
    \And
    Andrew D. Bagdanov \\
    andrew.bagdanov@unifi.it\\
    MICC, University of Florence}

\begin{document}

\maketitle

\begin{abstract}
Offline Reinforcement Learning (ORL) offers a robust solution to training agents in applications where interactions with the environment must be strictly limited due to cost, safety, or lack of accurate simulation environments. Despite its potential to facilitate deployment of artificial agents in the real world, Offline Reinforcement Learning typically requires very many demonstrations annotated with ground-truth rewards. Consequently, state-of-the-art ORL algorithms can be difficult or impossible to apply in data-scarce scenarios. In this paper we propose a simple but effective Reward Model that can estimate the reward signal from a very limited sample of environment transitions annotated with rewards. Once the reward signal is modeled, we use the Reward Model to impute rewards for a large sample of reward-free transitions, thus enabling the application of ORL techniques. We demonstrate the potential of our approach on several D4RL continuous locomotion tasks. Our results show that, using only 1\% of reward-labeled transitions from the original datasets, our learned reward model is able to impute rewards for the remaining 99\% of the transitions, from which performant agents can be learned using Offline Reinforcement Learning.
\end{abstract}

\section{Introduction}
\label{sec:submission}

Deep Reinforcement Learning \citep{DRL} is notoriously sample inefficient because agents can require hundreds of millions of interactions with the environment for convergence. This is prohibitive in many -- if not most -- real-world applications where interactions with the environment are expensive, dangerous, impossible or all of the above. The prohibitive sample complexity of Deep Reinforcement Learning is a barrier to its application in many scenarios where high-fidelity environment simulators are unavailable.

One of the most promising recent trends in Deep Reinforcement Learning aimed at mitigating the sample inefficiency of traditional approaches is Offline Reinforcement Learning~\citep{ORLLevine, ORLPrudencio}. Offline RL, sometimes also called Batch Reinforcement Learning, aims to learn policies from a fixed dataset of sampled agent-environment interactions, which turns the Reinforcement Learning problem into a more tractable \textit{supervised} learning problem. Datasets for Offline Reinforcement Learning typically contain state-action-reward-next state tuples, collected either from human experts interacting with the environment or from random exploration. However, Offline Reinforcement Learning algorithms assume that the total reward function is known and well-defined. This assumption may not be valid in many real-world scenarios, where the reward function may be noisy, sparse, or even missing, thus limiting the applicability of Offline RL.
To deal with the problem of missing reward, Imitation Learning \citep{ILsurvey, ILsurvey17} and Inverse Reinforcement Learning \citep{IRLRussell,IRLsurvey} can be used as alternative approaches, but their application introduces new constraints that must be handled appropriately.

Imitation learning allows an agent to learn a policy by observing and imitating the behavior of an expert, rather than relying solely on environmental rewards. This approach is particularly useful when the reward signal is unreliable, poor, or even absent, as it allows the agent to infer the correct actions simply by emulating the expert. The requirement for expert knowledge often cannot be met for realistic applications, limiting agents to less-than-optimal behavior.
Inverse Reinforcement Learning , on the other hand, takes a different approach to learning from expert behavior. Instead of directly imitating actions, like in Imitation Learning, IRL aims to infer the underlying reward function that the expert is implicitly optimizing. Once this reward function is estimated, standard Reinforcement Learning techniques can be used to find an optimal policy for the learned reward. 
Despite the ability of Imitation and Inverse Reinforcement Learning techniques to alleviate the missing reward problem, they implicitly assume the presence of optimal demonstrations or unlimited access to the environment.
Compared with existing solutions, the Reward Model we propose is free from all previous requirements: the reward signal is modeled from a mere distribution of reward-labeled transitions of unknown quality via supervised learning. Therefore, the environment is never accessed and no prior knowledge about the underlying distribution is required.

We focus on developing an effective technique to alleviate the limitations of Offline Reinforcement Learning in scenarios with highly unbalanced datasets -- that is, in environments in which the reward signal is defined for only a small portion of the available transitions.
Our reward model is a simple two-layer MLP which, when trained using only 1\% of the available transitions, can be used to impute rewards for the remaining 99\% of the dataset. This new dataset (with 99\% of the transitions using imputed rewards) is then used to perform Offline RL. 
This approach resulted in a consistent increase in ORL performance in terms of average D4RL scores for TD3BC \citep{TD3BC} and IQL \citep{IQL} agents, compared to applying offline RL in data sparsity scenarios.

The rest of the paper is organized as follows. In the next section, we discuss recent work from the literature most related to our contribution. In Section~\ref{sec:imputing} we introduce our reward modeling and reward imputation framework, and in Section~\ref{sec:experiments} we report results of experiments performed on the D4RL MuJoCo continuous locomotion tasks. We conclude in Section~\ref{sec:conclusion} with a discussion of our contribution.

\begin{figure}
    \begin{center}
        \includegraphics[width=1\linewidth]{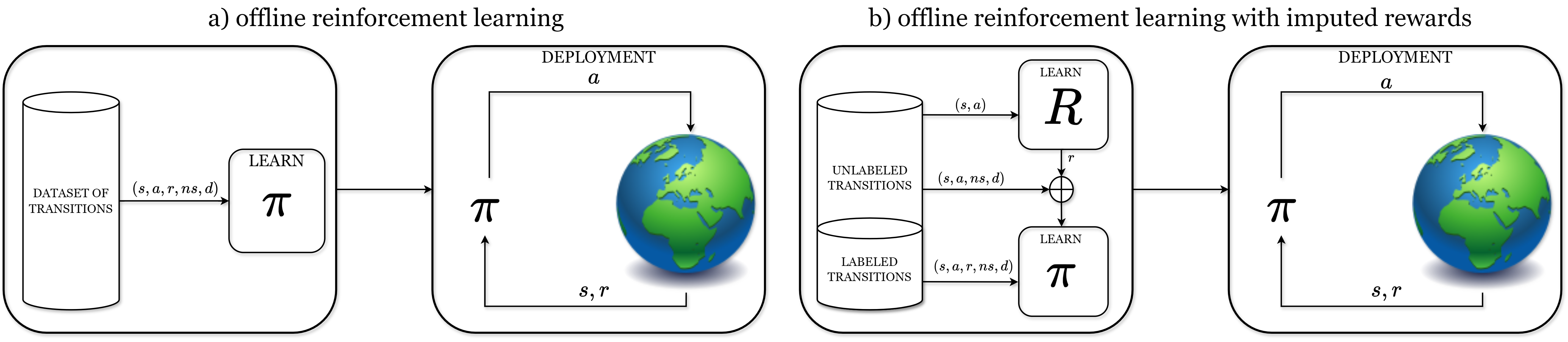}
    \end{center}
    \caption{Illustration of the scenarios of interest. (a) Classical offline reinforcement learning solutions are trained with a large set of transitions in which the reward signal is fully defined. On the other hand, in real applications, the reward signal may be available for only a small fraction of the total transitions, as in (b). In this case, the ORL algorithms are forced to use only those transitions where the reward signal is present, because of their inability to exploit the entire distribution, unless the reward signal is modeled from the distribution of reward-labeled transitions.}
    \label{fig:orl_scenarios}
\end{figure}

\section{Related Work}
\label{sec:related}

Behavior Cloning~\citep{BCPomerleau}, is based on a supervised learning approach and is the simplest Imitation Learning algorithm. Due to its way of modeling a policy, its application is limited to datasets of optimal (or near optimal) experiences. Moreover, a Behavior Cloning agent performs best when presented with observations close to the training set distribution; otherwise, its performance collapses \citep{SQIL}. A solution to this problem was proposed by DAgger ~\citep{DAGGER} in which the agent leverages its own induced state distribution to reduce the influence of out-of-distribution states. However, DAgger requires human intervention to relabel the input distribution. The high sample complexity and the requirement for expert demonstrations restrict the wide application approaches based on Behavior Cloning. In contrast, our solution enables Offline Reinforcement Learning by modeling the reward signal without requiring either the intervention of human expert operators or expert-quality transitions.

An alternative is offered by Inverse Reinforcement Learning \citep{IRLRussell} approaches in which an IRL algorithm, from the expert distribution of demonstrations, can model the underlying reward signal implicitly optimized, which is essential for policy learning through interactions with the environment. 
IRL agents potentially offer more solid performance than BC mimics by engaging the learning process from a Reinforcement Learning point of view. However, the learning process outlined by IRL agents has its own limitations. With respect to BC approaches, IRL agents are eager for computational resources due to their interactions with the environment. Moreover, the typical trial-and-error learning process of Reinforcement Learning algorithms can be dangerous or unfeasible to conduct in certain real-world scenarios. The sample complexity problem of IRL agents has been addressed by implementing generative approaches as in \citep{GAIL}. But once again, the presence of optimal/expert demonstrations is required. In our solution, we model the reward signal in an offline setting, thus avoiding any interaction with the environment or particular requirements for optimal/expert demonstrations.

Offline Inverse Reinforcement Learning tries to alleviate the above limitations. Offline IRL approaches can be roughly categorized into two branches: adversarial approaches \citep{OffIRL,ValueDice} that leverage GANs~\citep{GAN} to minimize the distance between the agent and the expert behaviors but suffer from vanishing gradient and convergence problems \citep{GANproblems}, and reward learning approaches \citep{rewardIRL,ORIL} that, on the other hand, imply running ORL solutions in an inner loop, thus increasing the overall complexity. An alternative approach is offered by \cite{DWBC} that, by contrasting expert with suboptimal demonstrations, add a weighting factor for the Behavioral Cloning term, but it requires extra effort to distinguish transitions based on their quality. The Reward Model we propose is able to impute rewards for reward-free transitions without the need to know the quality of the underlying distribution, thus addressing the modeling process only via label supervision.

\section{Imputing Rewards for Offline Reinforcement Learning}
\label{sec:imputing}

\subsection{Preliminaries}
\label{sec:preliminaries}

Reinforcement Learning~\citep{sutton1998introduction} aims to control a dynamic system, which is defined by a Markov Decision Process (MDP). The Markov Decision Process is defined by a tuple $\mathrm{M} = (S, A, T, d_0, r, \gamma)$: $S$ is the state space that could be continuous or discrete; $A$ is the action space that could also be continuous or discrete; $T$ represents the transition function that maps a \emph{(state, action)} pair at time $t$ into the state at next timestep, $T(s_{t+1}|s_t, a_t) \label{func:transition}$; $d_0$ is the initial state distribution $d_0(s_0)$ that indicates in what state the environment is initialized; $r$ is the reward function defined as $r:SxA \xrightarrow{} \mathrm{R}$ that represents the feedback signal for taking action $a$ in the state $s$; $\gamma \epsilon (0, 1]$ is the discount factor that weights short-term rewards with respect to long-term rewards.

The objective of Reinforcement Learning is to learn the parameters of an agent $\pi$ to maximize the expected future discounted reward $J(\pi)$ obtained by the agent in environment $M$:
\begin{equation}
    J(\pi) = \mathrm{E}_{\tau \sim p_\pi(\tau)}\left[\sum_{t=0}^{H}\gamma^tr(s_t, a_t)\right],
    \label{eq:cost-function}
\end{equation}
where $\gamma$ is a discount factor controlling the trade off between immediate and future reward. Reinforcement learning approaches typically require millions of interactions with the environment in order to minimize this objective $J$, given that the environment dynamics $T$ and reward signal $r$ are typically unknown.

Offline Reinforcement Learning is a data-driven approach to the RL problem. In this setting, the goal is still to optimize the objective function $J$, but without access to the environment. Instead, the agent is trained on static dataset of transitions $D = {(s_t^i, a_t^i, s_{t+1}^i, r_t^i)}_{i=1}^{N}$. An Offline RL algorithm should maximize $J$ having access only to this restricted training set of labeled environment transitions. In this work we consider scenarios in which the \textit{labeled} transitions available are very few, but we assume access to a large number of \textit{unlabeled} transitions (i.e. no instantaneous reward is available) on which we impute rewards.

\subsection{Methodology}

We consider Offline RL scenarios in which only a very small sample of reward-labeled transitions are available. That is, the training dataset consists of a labeled subset of $N_1$ transitions labeled with rewards:
\begin{eqnarray}
    \mathcal{D_l} = \{ (\mathbf{s}_n, \mathbf{a}_n, r_n) \}_{n=1}^{N_l},
\end{eqnarray}
and a much larger set of unlabeled transitions:
\begin{eqnarray}
D_u = \{ (\mathbf{s}_n, \mathbf{a}_n) \}_{n=1}^{N_u}
\end{eqnarray}

The Reward Model MLP $f(\mathbf{x}, \mathbf{a}; \theta)$ is trained using supervised learning on a dataset $\mathcal{D_l}$ containing state-action-reward triples to minimize the expected Mean Squared Error on the training set $\mathcal{D_l}$:
\begin{equation}
    \mathcal{L}(\mathcal{D_l}, \theta) = \mathbb{E}_{(\mathbf{x}, \mathbf{a}, r) \sim \mathbf{D_l}} \left[ (f(\mathbf{x}, \mathbf{a}; \theta) - r)^2 \right].
    \label{eq:rm-loss}
\end{equation}
We then use this trained Reward Model to impute missing rewards in $D_u$ and construct the Offline RL dataset with imputed rewards:
\begin{eqnarray}
D = D_l \cup \!\!\! \bigcup_{(\mathbf{x}, \mathbf{a}) \in D_u} \!\!\! \{ (\mathbf{x}, \mathbf{a}, f(\mathbf{x}, \mathbf{a}; \theta) \}
\end{eqnarray}
This dataset with imputed rewards can be used with any offline RL algorithm to train an agent. 
In the next section, to address data scarcity scenarios, we evaluate agents in which $D_l$ contains only 1\% of reward-labeled transitions.
Consequently, the final offline dataset $D$ consists of 99\% of rewards imputed via our Reward Model.

\section{Experimental Results}
\label{sec:experiments}

In this section we report on a range of experiments we performed to evaluate the effectiveness of imputing rewards for Offline Reinforcement Learning using out reward model.

\subsection{Environments and Datasets}
    We use D4RL~\citep{D4RL} continuous locomotion environments based on MuJoCo~\citep{MuJoco}. D4RL  provides offline RL datasets for these simulated environments, and have become standard benchmarks for Offline RL. We report results on the are Halfcheetah, Walker2D, and Hopper environments.
    For each environment we consider three variants of offline experience, each collected by a Reinforcement Learning agent at different points of its training::
\begin{itemize}
    \item \textbf{Medium-Replay}: this dataset consists of all experience the agent collected during its training until achieving Medium quality. This implies that it also contains random or near-random trajectories, corresponding to the initial steps of the training process. This dataset is typically difficult for Behavior Cloning approaches.
    \item \textbf{Medium}: This dataset is a collection of trajectories collected by an RL agent trained to almost half of its maximum ability.
    \item \textbf{Medium-Expert}: This dataset is a mixture of trajectories from the Medium and the Expert datasets.
\end{itemize}
    Non-expert quality distributions have mostly incomplete trajectories, with different degrees of exploration of the state-action space. As quality increases, highly rewarding regions are encountered more frequently.

\subsection{Training Details}

    Our Reward Model is trained to minimize  Equation~\ref{eq:rm-loss} for 300 epochs, using the Adam optimizer with a learning rate of $1e^{-4}$.
    To make a fair comparison, we used the parameters of both the original implementations of IQL and TD3BC.

\subsection{Baselines and Performance Evaluation}
    To quantify the performance of our solution, we compare the scores obtained by the IQL and TD3BC agents on different distributions.
    Each reported score represents the average of the normalized rewards, also called D4RL scores, collected during the evaluation in the last epoch of the offline training process.
    Each evaluation consists of ten episodes (1000 steps each) of online interactions with the environment.
    From each of these ten episodes, we take the normalized average of the last ten transitions. 
    The whole process is repeated for five random seeds.

\subsection{Comparative Performance Evaluation}

\begin{table}
    \scriptsize
    \centering
    \begin{tabular}{ll|rrr|rr|rr}
        \toprule
        & \multirow{2}[4]{*}{Environment} & \multicolumn{3}{c|}{Baseline (100\%)} & \multicolumn{2}{c|}{1\%} & \multicolumn{2}{c}{1\% + Imputed} \\
        \cmidrule{3-9}        
        & \multicolumn{1}{r|}{} & \multicolumn{1}{r}{TD3BC} & \multicolumn{1}{r}{IQL} & \multicolumn{1}{r|}{BC} & \multicolumn{1}{r}{TD3BC} & \multicolumn{1}{r|}{IQL} & \multicolumn{1}{r}{TD3BC} & \multicolumn{1}{r}{IQL} \\
        \midrule
        & halfcheetah-medium-v2 & 48.30 & 47.40 & 42.60 & 10.03 & 18.28 & 48.50 & 44.46 \\
        & hopper-medium-v2 & 59.30 & 66.30 & 52.90 & 11.32 & 15.53 & 58.13 & 36.47 \\
        & walker2d-medium-v2 & 83.70 & 78.30 & 75.30 & 6.00 & 1.21 & 82.81 & 69.87 \\
        & halfcheetah-medium-replay-v2 & 44.60 & 44.20 & 36.60 & 2.60 & 13.38 & 44.71 & 32.73 \\
        & hopper-medium-replay-v2 & 60.90 & 94.70 & 18.10 & 5.11 & 4.56 & 52.46 & 26.42 \\
        & walker2d-medium-replay-v2 & 81.80 & 73.90 & 26.00 & 5.37 & 1.44 & 69.15 & 47.38 \\
        & halfcheetah-medium-expert-v2 & 90.70 & 86.70 & 55.20 & 21.84 & 19.28 & 87.88 & 51.77 \\
        & hopper-medium-expert-v2 & 98.00 & 91.50 & 52.50 & 11.62 & 21.84 & 83.65 & 22.94 \\
        & walker2d-medium-expert-v2 & 110.10 & 109.60 & 107.50 & 8.77 & 8.04 & 89.30 & 107.38 \\
        \midrule
        & \textit{Gym locomotion-v2 average} & \textit{75.27} & \textit{76.96} & \textit{51.77} & \textit{9.19} & \textit{11.51} & \textit{68.51} & \textit{48.82} \\
        \midrule
    \end{tabular}
    
    \caption{The normalized results on D4RL Gym. \textit{Baseline} results are taken from the IQL paper~\citep{IQL}. The results for the \textit{1\%} and \textit{1\% + Imputed} scenarios are calculated by averaging mean returns over 10 evaluation trajectories and five random seeds. }
    \label{table:results}
\end{table}

    In Table~\ref{table:results}, we show the final performance obtained from agent training with the following settings:
    \begin{itemize}
        \item \textbf{Baseline}: the original datasets are used to measure the upper bounds;
        \item \textbf{1\%}: each ORL agent is trained only on transitions where the reward signal is available and well defined. In our scenario, this represents only 1\% of the transitions in the original dataset, previously called the train set;
        \item \textbf{1\% + Imputed}: each ORL agent is trained on the augmented dataset: from the previous scenario (1\%) the transitions where the reward signal was initially removed (99\%), but then reconstructed via our Reward Model, are combined.
    \end{itemize}

    From the column \textit{1\%}, we can see that, on average, the IQL algorithm performs better than TD3BC in the \textit{data scarcity} scenario. 
    Nonetheless, for both algorithms, the scores obtained by feeding the agents with such small datasets are far from the baselines.
    For Medium and Medium Replay distributions in the Halfcheetah environment with one percent of total transitions, IQL's application is more robust than TD3BC, although in the baseline scenario, TD3BC scores are better in that environment.
    In the Hopper environment, TD3BC is able to achieve better scores only in the Medium Replay distribution, while, on the other contrary, in the Walker2D environment, IQL fails.

    However, through the application of our Reward Model, the outcomes flip.
    Whenever in the 1\% scenario the TD3BC fails, now by accessing a wider distribution, this algorithm offers stronger performance, even better than the IQL counterpart.
    Catastrophic failures happen for the TD3BC algorithm with the Walker2D Medium and Halfcheetah Medium Replay distributions for the 1\% setting. In these particular distributions, we can see how the application of our solution strongly strengthens the final performance.
    For these distributions, the TD3BC agent is able to achieve almost the same performance as the baseline, or even better for the latter case.
    In addition, for the Halfcheetah Medium and Hopper Medium distributions, the TD3BC algorithm again achieves almost identical performance to the baselines.
    In the remaining distributions, the TD3BC algorithm achieves an average of 85 percent of baseline scores.

    The IQL algorithm through the expansion of the datasets, via imputed rewards, almost matches the baseline performance for the Halfcheetah Medium and Walker2D Medium Expert distributions.
    For the remaining environment-dataset combinations, IQL achieves around 56\% of baseline results.
    This is a solid improvement with respect to the \textit{1\%} scenario, where IQL is just able to achieve around 15\% of original scores.
    \begin{figure}
    \begin{center}
        \includegraphics[width=1\linewidth]{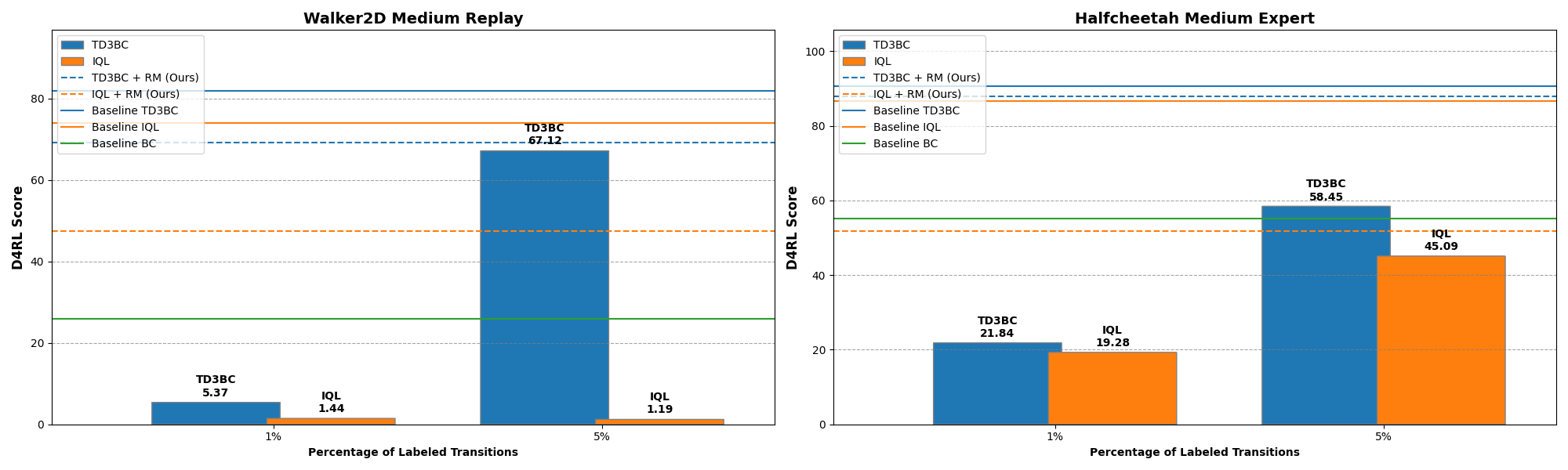}
    \end{center}
    \caption{Visual comparison of TD3BC and IQL results by changing the input distribution from 1\% to 5\% of reward-labeled transitions, for the Walker2D Medium Replay and Halfcheetah Medium Expert scenarios. In blue, we have the scores achieved by the TD3BC algorithm, whereas in orange, the IQL scores. The solid lines indicate the baseline scores of each algorithm trained on the original dataset. The dashed lines show the performance of our solution: only 1\% of reward-labeled transitions from the original dataset are considered, while the remaining 99\% of transitions are labeled by imputing the reward signal via our Reward Model. The solid green lines represent the score achieved by running BC agents on the original dataset.}
    \label{fig:barplot_hopper_medR}
\end{figure}

    In Figure~\ref{fig:barplot_hopper_medR}, we show a clear example of the real effectiveness of our solution. 
    We are comparing the D4RL scores achieved by TD3BC and IQL agents by varying the input distribution, from 1\% to 5\%. Moreover, we display the performance of the TD3BC and IQL agents trained on \textit{1\% + Imputed Reward} as dashed lines.
    On the left, we have the Walker2D Medium Replay case. For the smaller distribution (1\%), the final scores are unsatisfactory for both algorithms while, on the other hand, TD3BC (blue box) is able to exploit the larger distribution (5\%), thus achieving an almost-optimal score, even beating the BC (green solid line) baseline on the original dataset. However, our agent (dashed blue line) achieves a higher score.
    It is very important to note that to get closer to the performance of our solution, the TD3BC needs five times the number of input reward-labeled transitions.
    The IQL (orange box) algorithm trained on a larger distribution (5\%) collapsed, thus achieving lower results than the 1\% scenario. As the orange dashed line shows, our agent not only surpasses the IQL agent trained on the larger distribution but also it almost doubles the performance of the BC (green solid line) counterpart.  

    On the right, we analyze the Halfcheetah Medium Expert scenario. Once again, both algorithms suffer from data scarcity, thus achieving poor performance when the 1\% distribution is used. 
    By incrementing the number of input reward-labeled transitions, the TD3BC and IQL algorithms score higher results, but not sufficiently closer to the optimal results. 
    In particular, TD3BC (blue box) is showing better performance than the BC (green solid line) baseline, but it is way distant from our achievements (blue dashed line) which are almost optimal. IQL (orange box) trained on the broader distribution again performs worse than us (dashed orange line), but the scores obtained by both our agent and IQL are lower than agent BC (solid green line).
    We want to highlight that the optimal TD3BC score trained on the original dataset is 90.7 and, in this scenario, TD3BC trained on 5\% of transitions from the original dataset achieves 58.45.
    However, our solution by leveraging \textit{five times less reward-labeled transitions} scores 87.88.

\section{Conclusion}
\label{sec:conclusion}

In this paper, we have proposed a simple but effective solution for offline reinforcement learning in scenarios with extreme data sparsity in which the reward signal is incomplete. The reward model we propose is able to learn the reward signal from a very limited sample of transitions labeled with reward. Because of this, we can impute rewards for a large set of reward-free transitions, which most offline reinforcement learning algorithms are unable to exploit. The D4RL benchmark for MuJoCo tasks is used to compare our solution with TD3BC and IQL agents, where both algorithms represent the state-of-the-art for continuous control tasks. Considering an imperfect data scenario, such as having only 1\% of input reward-labeled transitions, our Reward Model is able to impute the reward signal for a wider distribution of reward-free transitions.
Reward signal imputation allows TD3BC and IQL agents to be trained on the entire distribution of offline experiences, which results in a substantial increase in performance, in terms of D4RL score, compared to simply applying ORL algorithms to data sparsity scenarios.


\bibliography{main}
\bibliographystyle{rlc}


\end{document}